\documentclass[letterpaper, 10 pt, conference]{ieeeconf}  

\IEEEoverridecommandlockouts                              

\usepackage[pdftex]{graphicx}
\usepackage{amsmath}
\usepackage{bm}
\usepackage{amssymb}
\usepackage{url}

\usepackage{tabularx}

\makeatletter
\let\NAT@parse\undefined
\makeatother

\usepackage{hyperref}
\hypersetup{
    colorlinks,
    linkcolor=magenta,
    filecolor=magenta,      
    urlcolor=magenta,
}
\def\secref#1{Sec.~\ref{#1}}
\def\figref#1{Fig.~\ref{#1}}
\def\tabref#1{Tab.~\ref{#1}}
\def\eqref#1{Eq.~(\ref{#1})}

\newcommand{\M}{SegLocNet}

\usepackage{multirow}
\usepackage{booktabs}
\usepackage{subcaption}
\usepackage{float}

\title{\LARGE \bf
\M: Multimodal Localization Network for Autonomous Driving via Bird’s-Eye-View Segmentation
}

\author{Zijie Zhou, Zhangshuo Qi, Luqi Cheng, and Guangming Xiong$^*$
\thanks{This work was supported by the National Natural Science Foundation of China under Grant 52372404.}%
\thanks{Zijie Zhou, Zhangshuo Qi, Luqi Cheng, and Guangming Xiong are with Beijing Institute of Technology, Beijing, 100081, China. * is the corresponding author.
        }%
\thanks{Our open-source code will be available at~\url{https://github.com/ZhouZijie77/SegLocNet}.}%
}

\begin{document}

\maketitle
\thispagestyle{empty}
\pagestyle{empty}

\begin{abstract}

Robust and accurate localization is critical for autonomous driving.
Traditional GNSS-based localization methods suffer from signal occlusion and multipath effects in urban environments.
Meanwhile, methods relying on high-definition (HD) maps are constrained by the high costs associated with the construction and maintenance of HD maps.
Standard-definition (SD) maps-based methods, on the other hand, often exhibit unsatisfactory performance or poor generalization ability due to overfitting.
To address these challenges, we propose~\M, a multimodal GNSS-free localization network that achieves precise localization using bird’s-eye-view (BEV) semantic segmentation.
\M\ employs a BEV segmentation network to generate semantic maps from multiple sensor inputs, followed by an exhaustive matching process to estimate the vehicle's ego pose.
This approach avoids the limitations of regression-based pose estimation and maintains high interpretability and generalization.
By introducing a unified map representation, our method can be applied to both HD and SD maps without any modifications to the network architecture, thereby balancing localization accuracy and area coverage.
Extensive experiments on the nuScenes and Argoverse datasets demonstrate that our method outperforms the current state-of-the-art methods, and that our method can accurately estimate the ego pose in urban environments without relying on GNSS, while maintaining strong generalization ability. Our code and pre-trained model will be released publicly.

\end{abstract}

\section{INTRODUCTION}

The field of autonomous driving has witnessed remarkable advancements over the past few decades, drawing continuous attention from both academia and industry. Within this domain, accurate and robust localization stands out as one of the most vital modules, essential for ensuring safe and stable autonomous driving. Global Navigation Satellite Systems (GNSS) can provide vehicles with global absolute positions. However, in urban environments, the presence of structures such as tunnels and buildings can obstruct signals and lead to multipath effects, resulting in inaccuracies in GNSS-based localization methods~\cite{zhang2023lidar, garg2021your}.

Consequently, achieving accurate localization under GNSS-denied conditions has always been a compelling area of research interest. Previous studies have utilized Simultaneous Localization and Mapping (SLAM) methods~\cite{zhang2014loam, mur2015orb, shan2018lego, xiong2023multi} for localization in such situations, but these approaches often entail high memory consumption and are not suitable for large-scale outdoor environments. 

In addition to SLAM, place recognition has also emerged as a potent solution to address this challenge~\cite{uy2018pointnetvlad,ma2022overlaptransformer,zhou2023lcpr}. By incorporating additional sensors and prior information, global relocalization can be achieved by the correlation of real-time observations with a comprehensive database of previously collected data.
Nonetheless, place recognition typically yields only approximate positions, with its accuracy being constrained by the data collection intervals within the database.

\begin{figure}
  \centering
  \includegraphics[width=1\linewidth]{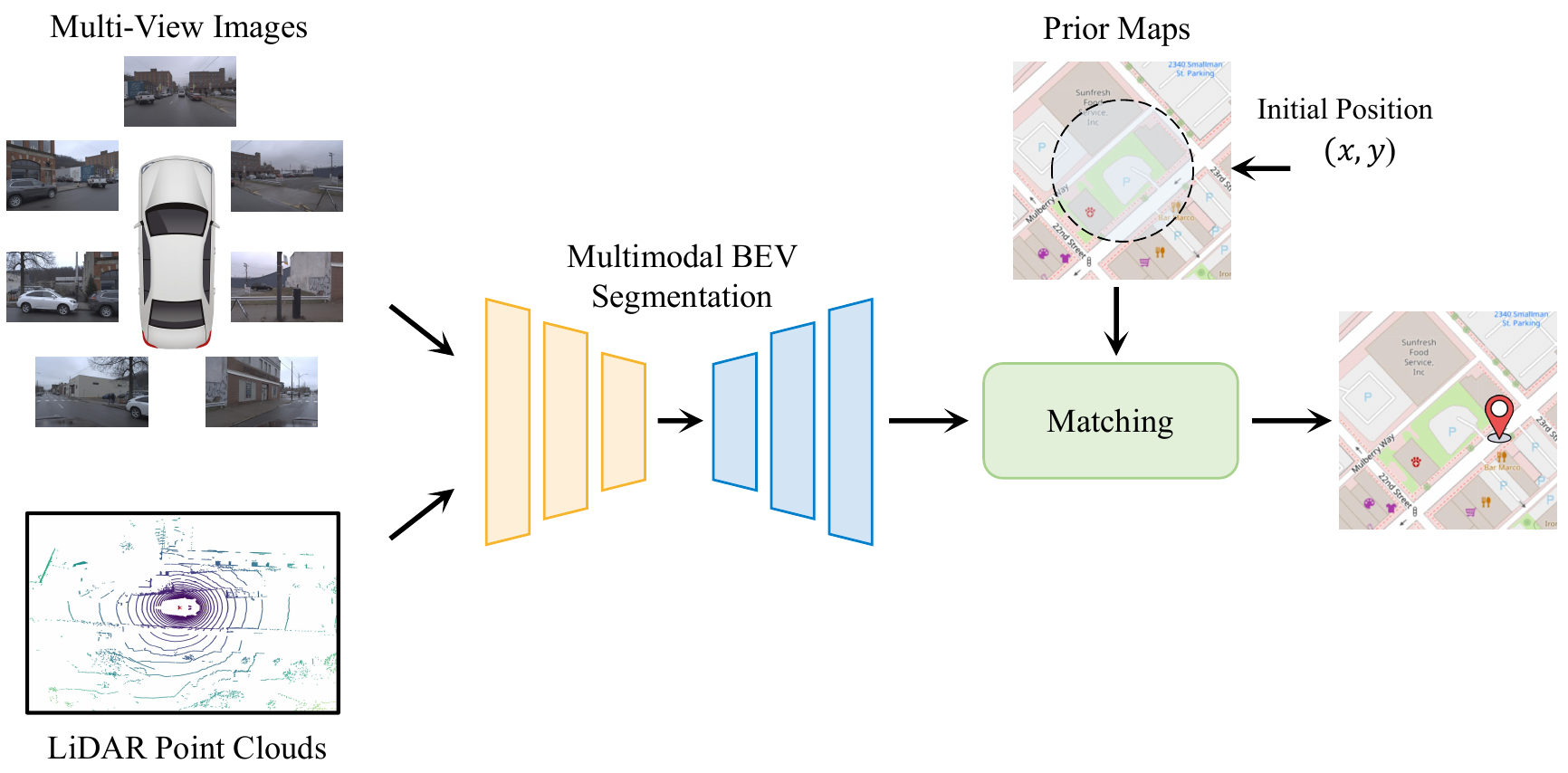}
  \caption{To address the localization challenges due to signal occlusion and multipath errors in urban environments, we propose~\M, a novel multimodal localization network that leverages multi-view images and the LiDAR point cloud to construct the local BEV semantic map of the surroundings. The precise pose estimation is achieved through the alignment of the BEV semantic map with the lightweight prior map.}
  \label{fig:intro}
  \vspace{-0.5cm}
\end{figure}

Recently, there has been a surge in interest in methods that rely on bird's-eye-view (BEV) perception and prior semantic maps~\cite{zhang2025bev, camiletto2024u, sarlin2023orienternet, wu2024maplocnet}.
Based on an intuition that human drivers can roughly recognize their locations by correlating what they see and the navigation maps, these approaches enable the direct estimation of the vehicle's ego pose from sensor inputs and semantic maps, circumventing the complexities associated with traditional 3D geometric feature matching processes.
Typically, these methods encode sensor inputs into an explicit or implicit local BEV representation via neural networks, which are then aligned with a learned or off-the-shelf semantic map based on an approximate prior position to obtain a more precise ego pose.

Despite the progress made in recent years, the performance of these methods remains unsatisfactory, and the localization mechanisms of some of the methods are still unexplainable, which hinders their generalization ability. To address these challenges, we propose~\M, a simple yet effective method that achieves high localization accuracy while maintaining strong generalization ability. 

As shown in~\figref{fig:intro}, our method accepts multi-view images and LiDAR point clouds as input, learns static elements in the environments through BEV segmentation task, and then matches the segmentation results with the prior map.
Notably, our method can be applied not only to a lightweight high-definition (HD) maps for higher performance but also to standard-definition (SD) maps such as OpenStreetMap (OSM)~\cite{osmpaper}, to cover a broader range of areas.

In summary, the main contributions of our work are as follows:
\begin{itemize}
\item We propose~\M, a multimodal localization network that achieves highly accurate localization in GPS-denied areas by incorporating the BEV segmentation and 2D navigation maps, while maintaining strong generalization ability.

\item We propose a binary map mask-based unified map representation. This design enables our method to be applied to both HD and SD maps without any modifications to the network architecture.

\item We conduct comprehensive experiments with detailed analysis to validate that our \M\  outperforms the current state-of-the-art methods.
\end{itemize}

\section{RELATED WORK}

\subsection{Localization Using Prior Maps}
An important method of localization in GNSS-denied environments is to match the pre-built maps to the sensor inputs. 
Several previous studies have employed feature matching techniques based on 3D maps to address this challenge~\cite{irschara2009structure, sattler2016efficient, toft2018semantic, sarlin2019coarse}. However, the creation and storage of 3D maps is expensive and their use in large-scale environments is difficult.
Consequently, recent research has focused on localization using more lightweight 2D maps.

Panphattarasap~\textit{et al.}~\cite{panphattarasap2018automated} present an image-based localization approach in urban environments by leveraging binary semantic descriptors (BSD) to match semantic features from images with the 2D map.
Subsequently, Yan~\textit{et al.}~\cite{yan2019global} advance this concept by incorporating the BSD descriptor into 3D LiDAR-based localization systems for mobile vehicles.
Samano~\textit{et al.}~\cite{samano2020you} design a siamese network to learn descriptors for both panoramic images and 2D map tiles in a low-dimensional embedded space.
Cho~\textit{et al.}~\cite{cho2022openstreetmap} propose a LiDAR-based vehicle localization method using OSM data instead of prior LiDAR maps. The localization is achieved by comparing the OSM descriptors and the LiDAR descriptors.
Based on the advantages of OSM being free and available globally~\cite{osmpaper}, Sarlin~\textit{et al.}~\cite{sarlin2023orienternet} propose OrienterNet, the first end-to-end image-to-OSM localization approach with high precision.

\subsection{BEV Representation for Autonomous Driving}
Learning powerful representations in BEV for perception tasks has been well studied in recent years.
LSS~\cite{philion2020lift} first introduces a well-known 2D-to-3D transformation framework based on depth. LSS predicts grid-wise depth distribution on 2D features, and then lifts the 2D features into 3D space via depth estimation.
Many subsequent works have adopted this approach~\cite{huang2021bevdet, li2023bevdepth, liu2023bevfusion}.
However, LSS-based methods are sensitive to the accuracy of depth estimation, some methods encode 2D features into 3D space via 3D-to-2D geometric projection, eliminating the need for explicit depth estimation~\cite{li2022bevformer, chen2022persformer}.
More recently, some studies further remove the 3D-to-2D geometric projection and instead implicitly learn the correlation between 2D features and BEV representations through the multi-layer perceptron (MLP) or transformer architecture~\cite{vaswani2017attention}.
For instance, HDMapNet~\cite{li2022hdmapnet} models the relationship between pixels in the perspective view and the camera coordinate system by an MLP.
CVT~\cite{zhou2022cross} uses a cross-view transformer to implicitly learn the geometric transformation. 
PETR~\cite{liu2022petr} introduces an innovative approach by integrating the 3D position embedding into image features, allowing the object queries to interact directly with 3D position-aware image features through vanilla cross attention, achieving a simpler and more elegant framework.

\subsection{Combining BEV Perception and Prior Maps}
Recently, there have been a series of works that combine BEV perception and prior maps to estimate vehicle's ego pose directly from sensor inputs and prior maps.
BEV-Locator~\cite{zhang2025bev} designs an end-to-end visual semantic localization network using multi-view camera images and first formulates the visual localization problem as an end-to-end learning scheme.
Building on a similar concept, OrienterNet~\cite{sarlin2023orienternet} achieves localization exclusively through monocular images and OSM, making it suitable for consumer-grade Augmented Reality (AR) applications.
Building upon OrienterNet, OSMLoc~\cite{liao2024osmloc} enhances the generalization ability by integrating the pre-trained foundation model~\cite{oquab2023dinov2} into the framework.
U-BEV~\cite{camiletto2024u} by Camiletto~\textit{et al.} employs a more powerful BEV representation based on the U-Net architecture~\cite{ronneberger2015u}, enabling localization in lower detailed maps.
He~\textit{et al.}~\cite{he2024egovm} propose a multimodal approach (EgoVM) that takes multi-view images, 3D LiDAR points, and offline vectorized maps to estimate the pose offset relative to the initial pose in an end-to-end manner.
MapLocNet~\cite{wu2024maplocnet} introduced by Wu~\textit{et al.} presents a coarse-to-fine registration framework for aligning BEV and neural map features, and investigates the impact of segmentation task on localization performance.

\begin{figure*}[ht]
  \centering
  \includegraphics[width=0.95\linewidth]{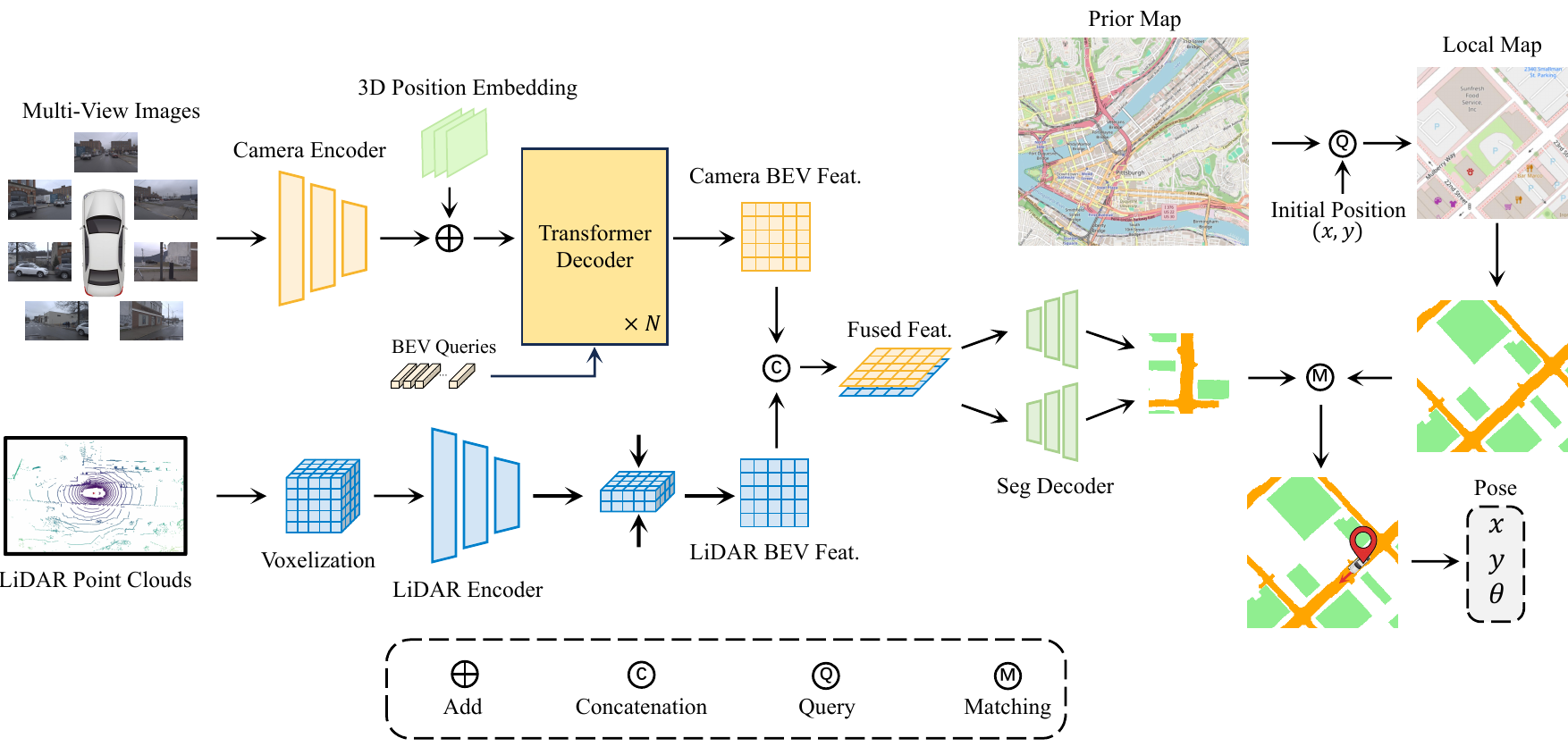}
  \caption{The overall architecture of~\M. We use a camera BEV encoder and a LiDAR BEV encoder to extract BEV features in a low-dimensional embedded space from the respective sensor inputs. Subsequently, the multimodal BEV features are fused with a lightweight convolutional neural network. The fused BEV features are then upsampled to the target resolution by the segmentation decoder to produce the BEV segmentation. Meanwhile, for each observation, a tile of local map is queried from the prior maps based on the given initial position. The vehicle's ego pose is estimated by exhaustively matching the BEV segmentation with the local prior map.}
  \label{fig:overall}
\end{figure*}

Inspired by the aforementioned works, we propose a novel multimodal end-to-end localization network that achieves high-accuracy localization performance with generalization by training a BEV segmentation network.
This design not only maximize the sharing of limited computational resources on the vehicle with other perception tasks, but also simplifies the integration of our method into an end-to-end autonomous driving framework~\cite{hu2022st, hu2023planning}.
Furthermore, by unifying the map representations of HD and SD maps, our method is capable of adapting to different types of maps with minimal changes.

\section{Our Approach}
\subsection{Overview}
\noindent\textbf{Problem Formulation.} Given synchronized multi-view images and LiDAR point clouds, along with a coarse initial position $(\check{x},\check{y})\in \mathbb{R}^2$, our goal is to estimate the precise position and orientation utilizing a lightweight prior map. Following the previous studies~\cite{sarlin2023orienternet, wu2024maplocnet, liao2024osmloc}, we simplify the standard 6-DoF pose estimation to a 3-DoF pose $\bm{p}=(x,y,\theta) \in \mathbb{R}^3$, which consists of the 2D position $(x,y)\in \mathbb{R}^2$ and the orientation angle $\theta \in (-\pi, \pi]$ around the $z$ axis. Here the $x$-$y$-$z$ axes correspond to east-north-up directions.
The initial position $(\check{x},\check{y})$ can be derived directly through GNSS, or through place recognition methods~\cite{uy2018pointnetvlad, zhou2023lcpr, ma2022overlaptransformer} in GNSS-denied environments.

\noindent\textbf{Overall Architecture.} The overall pipeline of our proposed method is illustrated in~\figref{fig:overall}, which mainly consists of four key components: the LiDAR BEV encoder, the camera BEV encoder, the semantic decoder, and the pose solver module.

Inspired by~\cite{liu2023bevfusion}, we first apply modality-specific encoders to extract features from sensor inputs across different modalities.
These features are subsequently encoded into the BEV space, providing a unified representation of the surrounding environment for localization.
Following this, multimodal feature fusion is performed within the BEV space, leading to a fused BEV feature that integrates both rich semantic and geometric information.
Subsequently, the semantic decoder is utilized to upsample the latent BEV features to the final output size and generate the segmentation results.

Finally, the pose solver module takes the semantic segmentation as input, matches it with the prior map that queried out with the given initial position, and identifies the position and the orientation that best aligns with the segmentation results to achieve 3-DoF localization. The above process can be expressed as
\begin{equation}
    \bm{p} =   f_\text{p}\left( \psi _\text{seg} \left(\mathcal{I}, \mathcal{P}\right), T_\text{init}\circ \mathcal{M}_\text{prior}  \right),
\end{equation}

where $\psi_\text{seg}(\cdot)$ represents the BEV segmentation network, $\mathcal{I}=\left \{ I_i \mid i=1,2,\cdots,N \right\}$ and $\mathcal{P}$ represent input multi-view images from $N$ views and the LiDAR point cloud respectively, $\mathcal{M}_\text{prior}$ represents the prior map, $T_\text{init}$ is the transformation derived from the initial position $(\check{x},\check{y})$, $\circ$ represents the spatial transform, and $f_\text{p}(\cdot)$ represents the pose solver module.
Further details of the components of our method are presented in the following sections.

\begin{figure}[t]
    \centering
    
    \begin{subfigure}[b]{0.32\linewidth}
        \centering
        \includegraphics[width=\linewidth]{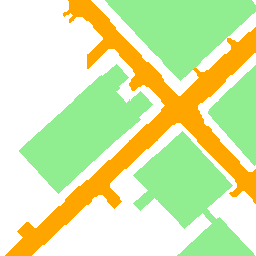}
        \caption{HD Binary Map}
    \end{subfigure}
    \hfill
    \begin{subfigure}[b]{0.32\linewidth}
        \centering
        \includegraphics[width=\linewidth]{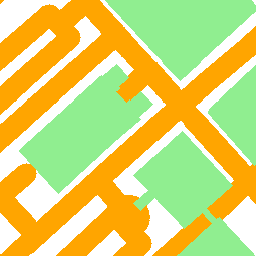}
        \caption{SD Binary Map}
    \end{subfigure}
    \hfill
    \begin{subfigure}[b]{0.32\linewidth}
        \centering
        \includegraphics[width=\linewidth]{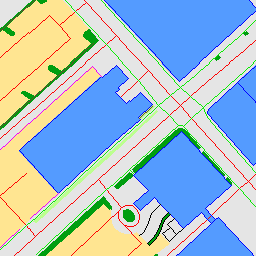}
        \caption{Rasterized Map}
    \end{subfigure}
    
    \vspace{1em}
    \begin{subfigure}[b]{\linewidth}
        \centering
        \includegraphics[width=\linewidth]{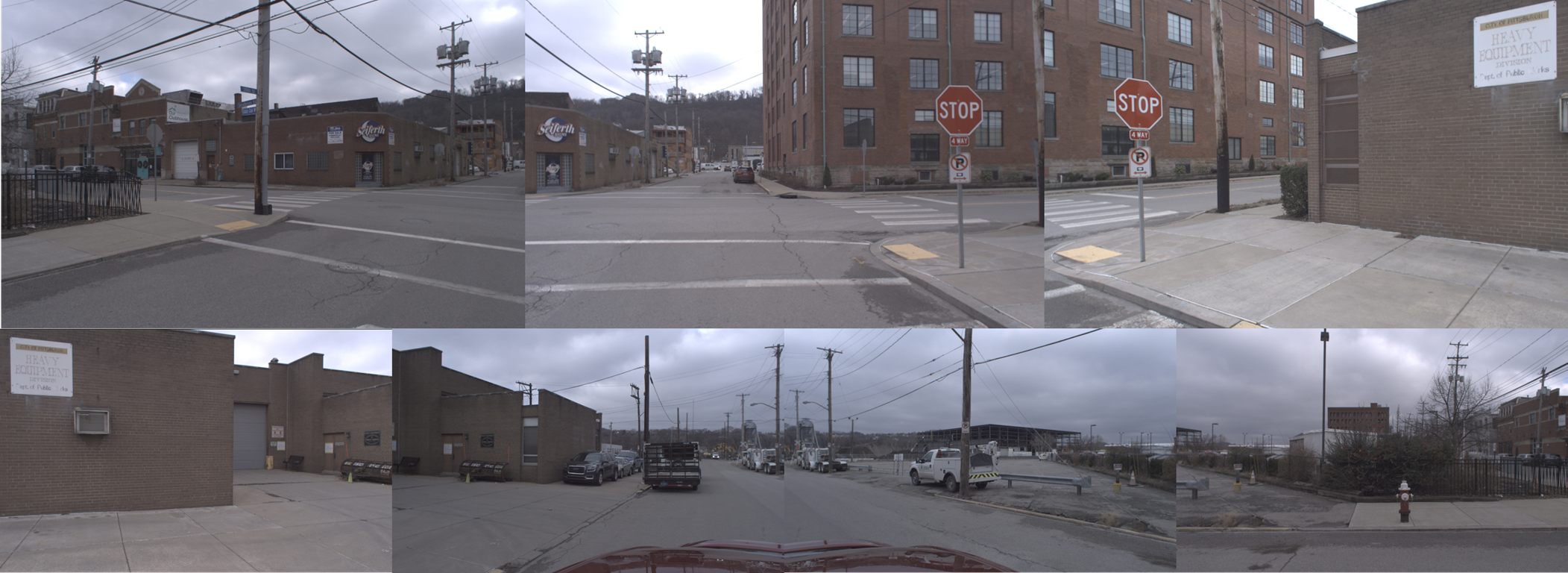}
        \caption{Surround-view Images}
    \end{subfigure}
    
    \caption{Visualization of different map processing methods.}
    \label{fig:mapshow}
    \vspace{-0.55cm}
\end{figure}

\subsection{Unified Map Representation}
\label{sec:mapform}
Our method integrates both HD and SD maps by employing a unified binary map mask representation for the prior map. This enables our method to adapt to different map forms with minimal adjustments.
In areas where HD maps are available, our method can produce more accurate results with the enhanced road information. Conversely, In areas without HD maps, SD maps can still be effectively utilized for localization.

\noindent\textbf{SD maps.}
Taking into account of the ease of accessibility and extensive coverage, we utilize OSM~\cite{osmpaper} as the data source for SD maps.
Previous studies~\cite{sarlin2023orienternet, wu2024maplocnet} categorize the OSM data into areas, ways, and nodes, and then rasterize these elements into 3-channel grid maps. 
However, considering the potential misalignment between OSM and sensor observations due to untimely updates, we extract only the most crucial elements for localization from OSM: roads and buildings. As a result, we obtain a 2-channel semantic binary map mask, serving both as the semantic labels for BEV segmentation and the prior map.
Moreover, since roads in OSM are represented as center-line skeletons, unlike previous methods that rasterize roads as polylines, we directly assign a certain width to roads during the rasterization process, which is more effective in the matching stage.

\noindent\textbf{HD maps.}
HD maps provide essential geometric and semantic information about road elements, while the construction and maintenance of HD maps are costly. 
In order to minimize our method's reliance on detailed road elements (such as \textit{lane divider}, \textit{road divider}, etc.), for HD maps, we utilize only the semantic labels of \textit{drivable area} as road information, while the building information is still derived from OSM.

Our processing for different map types are shown in~\figref{fig:mapshow}. For each observation, we query a binary map tile $F_m \in \mathbb{R}^{2 \times H_m \times W_m}$ centered around the initial position.

\subsection{BEV Segmentation}
\noindent\textbf{Camera Encoder.} Following the previous works~\cite{li2022bevformer,liu2022petr}, we adopt the transformer architecture for the camera BEV encoder.
Specifically, given the images $\mathcal{I}=\left \{ I_i \mid i=1,2,\cdots,N \right\}$ from $N$ views, we initially employ a 2D backbone (e.g. ResNet~\cite{he2016deep}, VoVNet~\cite{lee2020centermask}) to extract 2D image features $F^{2d} = \left \{ F^{2d}_i \in \mathbb{R}^{C\times H_F \times W_F} \mid i=1,2,\cdots, N \right \}$.
The 3D position embedding $p^{3d} = \left \{ p^{3d}_i \in \mathbb{R}^{C \times H_F \times W_F} \mid i=1,2,\cdots, N \right \}$ is derived from the 3D coordinates of the camera frustums, as detailed in~\cite{liu2022petr}.
Subsequently, the 3D position-aware image features $F^I \in \mathbb{R}^{C \times H_F \times W_F} $ are obtained by adding the 2D image features $F^{2d}$ and the 3D position embedding $p^{3d}$.
We initialize the BEV queries $F^I_\text{B}$ from the reference points sampled in the BEV space through a simple MLP.
The BEV queries are then fed into the transformer decoder to interact with the 3D position-aware image features through cross attention. For the transformer decoder, we use the same architecture as in DETR~\cite{carion2020end}.

\noindent\textbf{LiDAR Encoder.}
The raw LiDAR point cloud is first voxelized, and then fed into the LiDAR encoder based on the SECOND network~\cite{yan2018second} to extract LiDAR features, which is then flattened along the $z$-axis to generate the LiDAR BEV features.

We then fuse the generated image BEV features and the LiDAR BEV features through concatenation followed by a simple convolutional network. Considering computational consumption and real-time performance, we only employ a $1\times1$ convolution for feature fusion, which has yielded satisfactory results.

\noindent\textbf{Semantic Learning.}
We feed the fused BEV features into a fully convolutional semantic decoder, upsampling them $3$ times to the target resolution.
During the training phase, we use the semantic segmentation labels shown in~\figref{fig:mapshow} and apply sigmoid focal loss~\cite{ross2017focal} to supervise the two categories separately.
Finally, we directly take the semantic logits $F_\text{sem} \in \mathbb{R}^{2 \times H_\text{bev} \times W_\text{bev}}$ from the decoder as our segmentation results. 
In our approach, the difference between processing HD maps and SD maps lies in the learning of road information.
For HD maps, the network learns to identify drivable areas, which is consistent with the objectives of general BEV segmentation tasks. 
For SD maps, we manually specify the width of the roads in the semantic labels, prompting the network to focus more on the topological structure of the roads. This is why the network can converge as well, thus enabling the learning of HD maps and SD maps to share the same network architecture.

\begin{table*}[ht]
  \centering
  \begin{center}
  	\setlength{\tabcolsep}{2.8mm}
  	\renewcommand\arraystretch{1.25}
    \footnotesize{
        \begin{tabular}{cccccccccccc}
          \toprule
          \multirow{2}{*}{Methods} & \multirow{2}{*}{\#Cams$^1$} &\multicolumn{4}{c}{Position Recall@$X m \uparrow$}&\multicolumn{4}{c}{Orientation Recall@$X^{\circ} \uparrow$} & \multirow{2}{*}{APE($m$) $\downarrow$} & \multirow{2}{*}{AOE($^\circ$) $\downarrow$} \\ 
          \cmidrule(r){3-6} \cmidrule(r){7-10}
          ~ & ~ & $1m$ & $2m$ & $5m$ & $10m$ & $1^{\circ}$ & $2^{\circ}$ & $5^{\circ}$ & $10^{\circ}$ \\ \hline
          OrienterNet~\cite{sarlin2023orienternet} & S & 21.73 & 35.36 & 50.53 & 59.05 & 32.78 & 47.51 & 58.32 & 65.22 & 14.79 & 46.04 \\
          U-BEV*~\cite{camiletto2024u} & M & 16.89 & 41.60 & 71.33 & 81.46 & -- & -- & -- & -- & -- & -- \\
          MapLocNet*~\cite{wu2024maplocnet} & M & 20.10 & 45.54 & \underline{77.70} & \textbf{91.89} & \underline{58.61} & \underline{84.10} & \textbf{96.23} & \textbf{98.62} & -- & --\\
          \M-NM (ours) & M & 48.23 & 59.22 & 61.55 & 64.36 & 31.09 & 57.25 & 69.05 & 72.20 & 13.60 & 40.79 \\
          \M-SD (ours) & M & \underline{35.63} & \underline{57.98} & 74.55 & 80.09 & 38.58 & 64.98 & 83.62 & 87.40 & \underline{8.15} & \underline{19.68} \\
          \M-HD (ours) & M & \textbf{59.08} & \textbf{76.04} & \textbf{84.25} & \underline{86.86} & \textbf{63.19} & \textbf{84.55} & \underline{91.02} & \underline{93.00} & \textbf{5.30} & \textbf{10.11} \\ 
          \hline
          GT-HD & -- & 89.28 & 92.54 & 93.52 & 94.29 & 90.69 & 95.98 & 97.32 & 97.60 & 2.68 & 4.52 \\
          GT-SD & -- & 91.44 & 93.02 & 94.32 & 95.20 & 92.77 & 97.15 & 97.81 & 97.95 & 2.24 & 4.38 \\
          \bottomrule
        \multicolumn{12}{p{0.9\linewidth}}{$^1$ S: Single, M: Multiple}\\
        \end{tabular}
        }
    \caption{Location results on the nuScenes dataset. The sign * denotes the data is taken directly from the original paper. We use \textbf{bold} font to indicate the best results, and \underline{underline} to indicate the second-best results. $\uparrow$ means higher is better, $\downarrow$ means lower is better.}
    \label{tab:exp_nusc}
    \end{center}
    \vspace{-0.5cm}
\end{table*}

\subsection{Pose Solver} 
The key to achieving localization ultimately lies in how to effectively align the BEV perception results with the prior map.
MapLocNet~\cite{wu2024maplocnet} adopts a straightforward strategy: fusing BEV features with the neural map and then directly regressing the 3-DoF pose using MLPs.
However, in our experiments, we find that the regression-based method suffers from relatively slow convergence and fails to yield satisfactory performance, which is inconsistent with the results reported in their original paper.
A similar observation is also made in~\cite{liao2024osmloc}.
Moreover, the generalization capability of the regression-based prediction method is limited in areas without training data.
Therefore, we adopt the pose estimation method from OrienterNet~\cite{sarlin2023orienternet}. 
Specifically, we sample the semantic logits $F_\text{sem}$ at regular angular intervals for $K$ times, leading to candidate logits $F_\text{sem}' \in \mathbb{R}^{2 \times K \times H_\text{bev} \times W_\text{bev}}$. Subsequently, we exhaustively match $F_\text{sem}'$ with the binary map tile $F_m$ to obtain a score volume $M \in \mathbb{R}^{K \times H_m \times W_m}$ by
\begin{equation}
M_{k,h,w}=\sum_{c,i,j} {F'_\text{sem}}_{[ c,k,i,j] } \cdot {F_m}_{[c,h+i-1,w+j-1] }.
\end{equation}
$M_{k,h,w}$ denotes the matching score at the position $(h,w)$ on the prior map with the $k$-th sampling angle.
In practice, this exhaustive matching process can be accomplished by performing a convolution operation, where $F_\text{sem}'$ and $F_m$ serve as the input and the convolution kernel, respectively.

In addition, different from the previous studies, we do not use any neural networks to process the input maps. Experimental results demonstrate that directly matching the binary map mask and the semantic logits achieves satisfactory results. Detailed experiments are described in~\secref{sec:general}.

\begin{table}[!t]
  \centering
  \begin{center}
  	\setlength{\tabcolsep}{0.85mm}
  	\renewcommand\arraystretch{1.25}
    \footnotesize{
        \begin{tabular}{ccccccccc}
          \toprule
          \multirow{2}{*}{Methods} &\multicolumn{4}{c}{Position R@$X m \uparrow$}&\multicolumn{4}{c}{Orientation R@$X^{\circ} \uparrow$}\\ 
          \cmidrule(r){2-5} \cmidrule(r){6-9}
          ~ & $1m$ & $2m$ & $5m$ & $10m$ & $1^{\circ}$ & $2^{\circ}$ & $5^{\circ}$ & $10^{\circ}$ \\ \hline
          OrienterNet~\cite{sarlin2023orienternet} &  16.84 & 26.15 & 40.16 & 48.84 & 33.08 & 45.70 & 51.65 & 57.91 \\
          MapLocNet*~\cite{wu2024maplocnet} & 23.26 & 47.24 & 79.13 & \textbf{94.33} & \underline{62.35} & \underline{86.28} & \textbf{96.24} & \textbf{98.61} \\
          \M-SD  & \underline{46.97} & \underline{74.01} & \underline{81.23} & 84.93 & 51.77 & 73.87 & 85.46 & 85.84 \\
          \M-HD & \textbf{68.37} & \textbf{83.85} & \textbf{90.23} & \underline{92.01} & \textbf{63.43} & \textbf{86.92} & \underline{95.05} & \underline{95.05} \\
          \bottomrule
        \end{tabular}
        }
    \caption{Localization results on the Argoverse dataset.}
    \label{tab:exp_ftarg}
    \end{center}
    \vspace{-0.5cm}
\end{table}

\begin{table*}[ht]
  \centering
  \begin{center}
  	\setlength{\tabcolsep}{2.5mm}
  	\renewcommand\arraystretch{1.25}
    \footnotesize{
        \begin{tabular}{c|ccccccccccc}
          \toprule
          \multirow{2}{*}{Location} & \multirow{2}{*}{Methods} &\multicolumn{4}{c}{Position Recall@$X m \uparrow$}&\multicolumn{4}{c}{Orientation Recall@$X^{\circ} \uparrow$} & \multirow{2}{*}{APE($m$) $\downarrow$} & \multirow{2}{*}{AOE($^\circ$) $\downarrow$}\\ 
          \cline{3-6} \cline{7-10}
          ~ & ~ & $1m$ & $2m$ & $5m$ & $10m$ & $1^{\circ}$ & $2^{\circ}$ & $5^{\circ}$ & $10^{\circ}$ \\ \hline
          \multirow{5}{*}{Miami} & OrienterNet~\cite{sarlin2023orienternet} & 8.60 & 11.01 & 20.27 & 31.83 & 14.91 & 19.16 & 24.22 & 36.83 & 24.52 & 107.79 \\
          ~ & MapLocNet One-Stage~\cite{wu2024maplocnet} & 1.35 & 4.25 & 16.71 & 39.43 & 17.61 & 29.82  & 42.94 & 45.94 & 16.64 & 86.43 \\
          ~ & \M-NM (ours) & 43.69 & 61.66 & 73.87 & 78.32 & 45.39 & 73.52  & 79.77 & 80.28 & 8.52 & 40.22 \\
          ~ & \M-SD (ours) & \underline{48.44} & \underline{71.27} & \underline{80.28} &  \underline{86.23} & \underline{46.34} & \underline{74.47}  & \underline{82.28} & \underline{83.58} & \underline{6.22} & \underline{33.19} \\
          ~ & \M-HD (ours) & \textbf{55.90} & \textbf{75.97} & \textbf{85.88} & \textbf{87.83} & \textbf{56.45} & \textbf{82.88} & \textbf{93.29} & \textbf{93.64} & \textbf{4.26} & \textbf{14.85} \\         
          \hline
        \multirow{5}{*}{Pittsburgh} & OrienterNet~\cite{sarlin2023orienternet} & 0.27 & 0.92 & 3.50 & 9.72 & 1.93 & 3.73 & 6.86 & 11.29 & 33.34 & 110.20\\
          ~ & MapLocNet One-Stage~\cite{wu2024maplocnet} & 0.00 & 0.36 & 3.22 & 10.69 & 0.36 & 1.01  & 1.98 & 3.31 & 26.66 & 94.57 \\
          ~ & \M-NM (ours) & 9.58 & 21.10 & 27.60 & 30.18 & 17.83 & 29.07  & 46.82 & 50.23 & 25.39 & 68.29 \\
          ~ & \M-SD (ours) & \underline{32.99} & \underline{64.70} & \underline{73.68} & \underline{75.48} & \underline{39.58} & \underline{68.57} & \underline{78.47} & \underline{79.26} & \underline{11.09} & \underline{34.86} \\
          ~ & \M-HD (ours) & \textbf{55.71} & \textbf{76.63} & \textbf{83.91} & \textbf{86.17} & \textbf{57.18} & \textbf{82.02}  & \textbf{88.11} & \textbf{88.29} & \textbf{5.85} & \textbf{21.94} \\
          \bottomrule
        \end{tabular}
        }
    \caption{Generalization evaluation on the Argoverse dataset.}
    \label{tab:exp_miapit}
    \end{center}
    \vspace{-0.2cm}
\end{table*}

\begin{figure*}
    \centering
    \includegraphics[width=\linewidth]{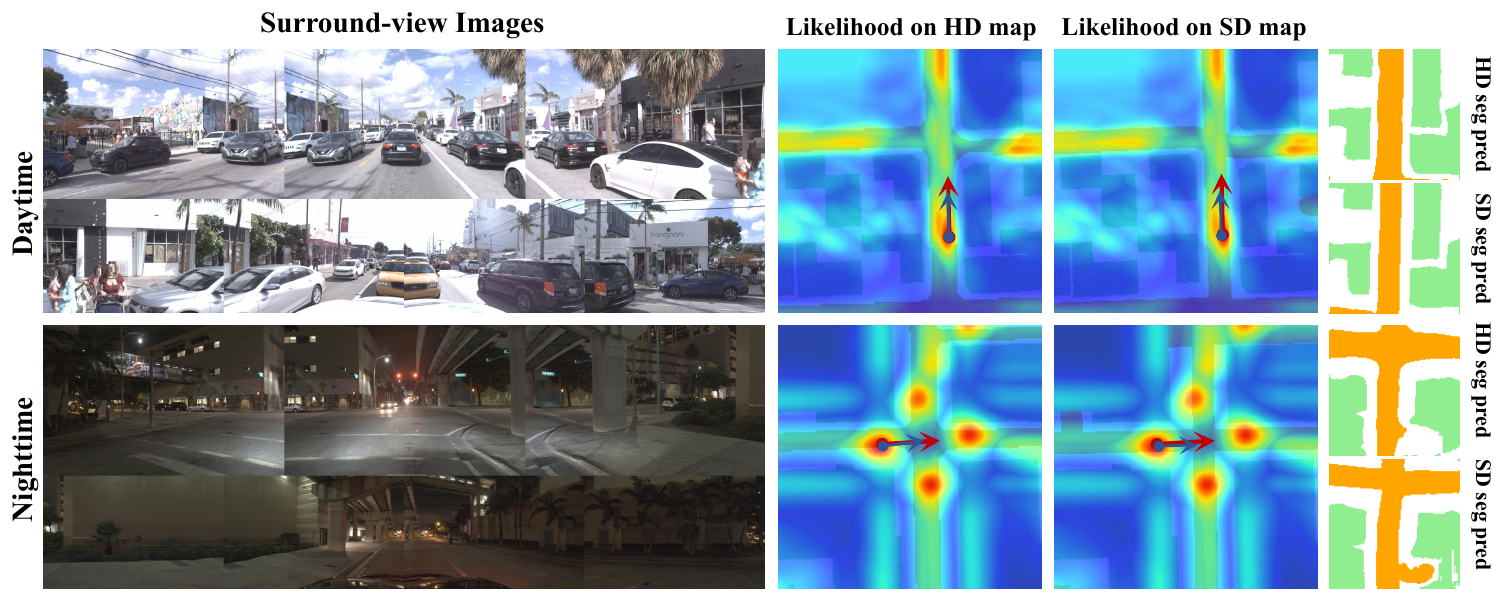}
    \caption{Visualization of localization results on the Argoverse dataset. The red and blue arrows represent the GT pose and the predicted pose respectively.}
    \label{fig:vis}
    \vspace{-0.2cm}
\end{figure*}

\section{EXPERIMENTS}
\subsection{Experimental Settings}
\noindent\textbf{Dataset.}
We train and evaluate our method on two autonomous driving datasets: nuScenes~\cite{caesar2020nuscenes} and Argoverse~\cite{chang2019argoverse}.
The nuScenes dataset contains 1000 driving scenes collected from two cities: Boston and Singapore. We use the standard data split, with 28130 samples for training, 6019 samples for validation, and 6008 samples for testing.
The Argoverse dataset contains 113 driving scenes collected from Miami and Pittsburgh, where the training set contains 13122 samples, the validation set contains 5015 samples, and the test set contains 4168 samples. 
Both datasets provide multi-view images, LiDAR point clouds, and HD maps. We supplement them with SD maps from OSM. However, the pose information provided by the nuScenes dataset could not be perfectly aligned with the OSM, so we follow~\cite{wu2024blos} to align them as much as possible by adjusting transformation parameters. The Argoverse dataset does not require this alignment procedure.

\noindent\textbf{Implementation Details.}
We employ VoVNetV2~\cite{lee2020centermask} as the image encoder and SECOND~\cite{yan2018second} as the LiDAR encoder. The input images are resized to a resolution of $320 \times 800$, while the input point clouds are voxelized with $0.5m$.
The perception range is defined as $(-32m, 32m)$, centered around the ego vehicle, and the size of the BEV grid map is set to $128 \times 128$.
To simulate the coarse initial position, for each frame, we randomly select a position within a range of $(-32m, 32m)$ around the ground truth position of the ego vehicle and extract a map patch of size $256 \times 256$ centered at this position.
Both the BEV grid map and the map patch are set to a resolution of $0.5$ meters per pixel.
We rotate the semantic logits $256$ times in the matching process, that is $K=256$.
During training, we optimize the network using AdamW~\cite{loshchilov2017decoupled}, with an initial learning rate of $1 \times 10^{-4}$ and a cosine annealing strategy. We train our model for $50$ epochs on 4 NVIDIA RTX 4090 GPUs.

\noindent\textbf{Baseline Methods.}
We compare our proposed method with OrienterNet\cite{sarlin2023orienternet}, MapLocNet\cite{wu2024maplocnet}, U-BEV\cite{camiletto2024u}, and three variants of our method.

\textit{a) OrienterNet:} OrienterNet is currently the only open-source comparable method in this domain. We use the officially released implementation and retrain it on the nuScenes and Argoverse datasets.

\textit{b) U-BEV:} Following~\cite{wu2024maplocnet}, we also report the results of U-BEV directly from their original paper. It should be noted that the output of U-BEV does not include orientation results, so we only report the 2-DoF pose prediction results.

\textit{c) MapLocNet:} MapLocNet's task is the closest to ours. We attempted to implement it according to their original paper, but only the one-stage version of MapLocNet could converge. Therefore, we report the results from both their original paper and our retrained one-stage version. The one-stage version is used for generalization evaluation to illustrate the adverse impact of regression-based pose estimation on generalization ability.

\textit{d) \M-SD/HD:}  We train and evaluate our proposed method with the SD maps and HD maps described in~\secref{sec:mapform}. The only difference between~\M-SD and~\M-HD lies in the form of the input maps used for localization.

\textit{e) \M-NM:} In this variant, we employ a U-Net architecture to process rasterized SD maps and generate neural maps, which are subsequently aligned with the BEV features. Similar to~\cite{sarlin2023orienternet}, we utilize the cross entropy loss to maximize the log-likelihood of the ground-truth (GT) pose on the score volume $M$.

\begin{table*}[ht]
  \centering
  \begin{center}
  	\setlength{\tabcolsep}{3.5mm}
  	\renewcommand\arraystretch{1.25}
    \footnotesize{
        \begin{tabular}{cccccccccccc}
          \toprule
          \multirow{2}{*}{Drivable} & \multirow{2}{*}{Building} &\multicolumn{4}{c}{Position Recall@$X m \uparrow$}&\multicolumn{4}{c}{Orientation Recall@$X^{\circ} \uparrow$} & \multirow{2}{*}{APE($m$) $\downarrow$} & \multirow{2}{*}{AOE($^\circ$) $\downarrow$} \\ 
          \cmidrule(r){3-6} \cmidrule(r){7-10}
          ~ & ~ & $1m$ & $2m$ & $5m$ & $10m$ & $1^{\circ}$ & $2^{\circ}$ & $5^{\circ}$ & $10^{\circ}$ \\ \hline
          ~ & \checkmark & 37.43 & 46.94 & 68.66 & 73.52 & 26.67 & 46.74 & 72.27 & 74.12 & 10.96 & 39.72 \\
          \checkmark & ~ & 42.54 & 72.92 & 85.08 & \textbf{88.58} & 50.35 & 77.12 & 86.83 & 86.93 & 5.05 & 23.89 \\
          \checkmark & \checkmark & \textbf{55.90} & \textbf{75.97} & \textbf{85.88} & 87.83 & \textbf{56.45} & \textbf{82.88} & \textbf{93.29} & \textbf{93.64} & \textbf{4.26} & \textbf{14.85} \\
          \bottomrule
        \end{tabular}
        }
    \caption{Ablation study on different map elements.}
    \label{tab:exp_ablation1}
    \end{center}
    \vspace{-0.5cm}
\end{table*}

\subsection{Evaluation for Localization}
\noindent\textbf{nuScenes.} We first evaluate our method on the nuScenes dataset, adopting the same evaluation metrics as those employed in the prior works~\cite{wu2024maplocnet, liao2024osmloc}, including Recall@$Xm$, Recall@$X^\circ$, Absolute Position Error (APE), and Absolute Orientation Error (AOE).
The experimental results are shown in~\tabref{tab:exp_nusc}.
As can be seen, our method significantly outperforms the baseline methods in terms of Recall@$1/2/5m$ and Recall@$1^\circ$, and also demonstrates strong performance on other metrics.
However, we observe that our results on Recall@$10m$, Recall@$5^\circ/10^\circ$ are slightly lower than those reported in MapLocNet.
We believe this discrepancy arises because MapLocNet directly regresses the offset from the initial pose to the GT pose, which may lead to overfitting with extensive training.
Additionally, our retrained OrienterNet also exhibits a large difference from the results presented in MapLocNet. Therefore, we consider the results reported in MapLocNet as a reference only and have marked them with * in the table.
Furthermore, we conduct an exhaustive matching experiment by directly aligning the prior maps with the GT BEV segmentation labels. The results are presented in the last two rows of~\tabref{tab:exp_nusc}.
It can be observed that high performance is achievable if the BEV segmentation is accurate enough, which underscores our approach of directly achieving localization through the training of a BEV segmentation model.

\noindent\textbf{Argoverse.} We further conduct experiments on the Argoverse dataset.
We fine-tune the model, pre-trained on the nuScenes dataset, for one epoch on the Argoverse dataset.
It is worth noting that MapLocNet does not specify the number of fine-tuning epochs, however we still include their results here for comparison purposes.
The experimental results are summarized in~\tabref{tab:exp_ftarg}.
Our method continues to demonstrate superior performance compared to OrienterNet across all metrics.
Although it still falls slightly short of the results reported in MapLocNet on Recall@$10m$ and Recall@$5^\circ/10^\circ$, it significantly outperforms it on other metrics, especially Recall@$1m$, showing the ability of our method to rapidly adapt to new environments with minimal training.
Moreover, we can also see that in areas that provide HD maps, the use of HD maps can achieve higher localization accuracy.
However, in areas lacking HD maps, due to the design of the unified map representation, our method can still achieve competitive performance with SD maps.

To intuitively demonstrate the performance of our method,~\figref{fig:vis} provides the visualization of the BEV segmentation and the likelihood maps during pose matching process.
We also showcase a nighttime scenario. Leveraging the advantages of multimodal fusion, our proposed method is still able to accurately estimate the vehicle's ego pose even in nighttime conditions.

\subsection{Generalization Ability}
\label{sec:general}
To investigate the generalization ability of our proposed method, we train our model on the Miami (MIA) split of the Argoverse dataset and directly test it on the Pittsburgh (PIT) split without any fine-tuning.
The results are shown in~\tabref{tab:exp_miapit}.
As can be seen, our proposed method significantly outperforms the retrained OrienterNet and MapLocNet One-Stage on the MIA split.
Additionally, without any fine-tuning, our method achieves remarkable performance on the PIT split, demonstrating its superior generalization ability.
In contrast, both OrienterNet and MapLocNet exhibit relatively weaker generalization capabilities.
Furthermore, we compare a variant of our method that utilizes neural maps extracted from prior maps, which also shows poor generalization performance.
Collectively, our experimental results suggest that both the extraction of neural maps from prior maps and the use of regression-based pose estimation negatively impact the model's generalization ability.
This is also why we do not adopt these two approaches, but instead directly use the BEV segmentation labels as the prior map and estimate the pose through an exhaustive matching method. In this way, we achieve a simple and elegant localization framework while maintaining excellent generalization ability.

\subsection{Ablation Studies and Analysis}
\label{sec:ablation}
In this section, we present the ablation study and analysis on the MIA split of the Argoverse dataset.
First, we perform ablation experiments to examine the contributions of different map elements, and the results are shown in~\tabref{tab:exp_ablation1}.
The results demonstrate that both the two map elements: \textit{drivable area} and \textit{building} positively contribute to localization accuracy.
Moreover, using only the element \textit{drivable area} achieves better performance than using only the element \textit{building}, suggesting that road structure has a more significant impact on localization.
Second, we investigate the influence of varying road widths during the construction of SD maps through experiments, with the experimental results detailed in~\tabref{tab:exp_ablation2}.
Although the road width of $2.5 m$ achieves the best Recall@$1m$ and Recall@$1^\circ$, a comprehensive evaluation of all other metrics indicates that the width of $10m$ provides a more balanced and superior overall performance.
Therefore, we select the $10m$ road width for our final implementation.

\begin{table}[t]
  \centering
  \begin{center}
  	\setlength{\tabcolsep}{0.85mm}
  	\renewcommand\arraystretch{1.25}
    
    \footnotesize{
        \begin{tabular}{>{\centering\arraybackslash}p{0.9cm}cccccccc}
          \toprule
          \multirow{2}{*}{Width} &\multicolumn{3}{c}{Position R@$X m\uparrow$}&\multicolumn{3}{c}{Orientation R@$X^{\circ}\uparrow$} & \multirow{2}{*}{APE($m$)$\downarrow$} & \multirow{2}{*}{AOE($^\circ$)$\downarrow$}\\ 
          \cmidrule(r){2-4} \cmidrule(r){5-7}
          ~ & $1m$ & $5m$ & $10m$ & $1^{\circ}$ & $5^{\circ}$ & $10^{\circ}$ \\ \hline
          2.5$m$ &  \textbf{53.55} & 75.17 & 78.12 & \textbf{62.86} & 80.68 & 81.58 & 8.81 & 36.90 \\
          5$m$ & 46.49 & 76.87 & 82.83 & 57.80 & 81.48 & 81.88 & 7.77 & 36.58 \\
          10$m$  & 48.44 & \textbf{80.28} & \textbf{86.23} & 46.34 & \textbf{82.28} & \textbf{83.58} & \textbf{6.22} & \textbf{33.19} \\
          \bottomrule
        \end{tabular}
        }
    \caption{Localization results with different road widths during the rasterization of SD maps.}
    \label{tab:exp_ablation2}
    \end{center}
    \vspace{-0.5cm}
\end{table}

\section{CONCLUSIONS}
In this work, we present~\M, a multimodal end-to-end localization nerual network, that utilizes BEV segmentation to achieve accurate and robust localization in complex urban environments.
Our method relies solely on the BEV segmentation task for training, which facilitates efficient sharing of limited computational resources on the vehicle with other perception tasks.
On the other hand, by introducing a unified map representation, our method can be applied to both HD maps and SD maps without any modifications to the network architecture.
Experimental results show that our method outperforms the current state-of-the-art methods on localization accuracy while maintaining strong generalization ability.










\bibliographystyle{IEEEtran}

\bibliography{IEEEabrv,ref}

\end{document}